\documentclass[runningheads]{llncs}
\usepackage[T1]{fontenc}
\usepackage{graphicx}
\usepackage{amsfonts}
\usepackage{array}
\usepackage{hyperref}
\usepackage[usenames]{color}

\usepackage{multirow}
\usepackage{pdflscape}   
\usepackage{array}
\usepackage{rotating}

\usepackage{graphicx}
\usepackage{subcaption}
\usepackage{booktabs}

\newcommand{\red}[1]{\textcolor{red}{#1}}

\usepackage{multirow}

\begin{document}
\title{Comparative Study of Domain-adapted VLMs \\for General Document Visual Question Answering}
    \titlerunning{Comparative Study of Domain-adapted VLMs for General DocVQA}
%
\author{Miguel Lopez-Duran \and 
Elena Marrero \and 
Julian Fierrez \and 
Marta Robledo-Moreno \and 
Ruben Vera-Rodriguez \and 
Daniel DeAlcala \and 
Aythami Morales \and 
Ruben Tolosana \and 
Oscar Delgado \and 
Alvaro Ortigosa \and 
Javier Ortega-Garcia} 
\authorrunning{M. Lopez-Duran, E. Marrero, J. Fierrez, et al.}
%
\institute{BiometricsAI,  Universidad Autónoma de Madrid (UAM), Spain \\
\email{miguel.lopezd@uam.es, julian.fierrez@uam.es}}
\maketitle              
\begin{abstract}

Document Visual Question Answering (DocVQA) presents a complex multimodal challenge, requiring models to exploit visual, textual, and layout information from documents. Although Vision-Language Models (VLMs) have shown remarkable performance in text-vision tasks, their robustness and transferability to different document domains remains underexplored. In this study, we present a comprehensive evaluation of $8$ open-source pretrained VLMs on DocVQA in three different document domains: industrial documents of varying type, infographics, and presentation slides. We systematically assess model performance under zero-shot evaluations, fully supervised finetuning with inter- and intra-dataset evaluations, and few-shot learning evaluations of knowledge transfer between domains. Our findings demonstrate that while large pretrained VLMs possess strong zero-shot baselines for structured layouts, their performance strongly decreases on visually complex layouts of infographics and slides. Although parameter scaling is a dominant factor on performance, supervised finetuning yields higher relative gains in smaller architectures. Furthermore, our cross-domain and few-shot experiments show that visual understanding is the main bottleneck for DocVQA, not a lack of knowledge from the VLMs. Using $50$ target domain samples, the models finetuned in DocVQA with datasets of different domains rapidly adapt to the target domain documents, even surpassing their fully supervised counterparts in some cases.

.
\end{abstract}
\section{Introduction}

Visual Question Answering (VQA) is a core multi-modal task in machine learning consisting of answering text-based questions about an image. Generally speaking, VQA tasks can be classified as extractive and abstractive Question Answering (QA). Extractive QA consists of answering the question by extracting a subset of tokens within the image. On the other hand, abstractive QA aims to answer the question based on the image content, but the answer does not need to be extracted from it. 

There are different VQA tasks depending on the application scenario, such as maps~\cite{chang2022mapqa}, daily photos~\cite{hudson2019gqa}, or scientific papers~\cite{pramanick2024spiqa}. Among all these tasks, Document VQA (DocVQA), which aims to answer questions based on document images, is much more challenging. It requires detecting the layout objects \cite{2023_ICDAR_DocumentLayout_Pena,PENA-Layout,lopez2025benchmarking} and extracting the relationships between them to extract relevant information to answer the question correctly. 

This complexity is even greater when different document domains are taken into account. Different document domains differ greatly from each other. For example, scientific papers usually present a structured layout with one or two columns, same fonts for all the objects, and the relations between objects are simple and much more direct. However, other document types, like presentation slides, do not have a structured layout and the relations between objects are less structured in order to be visually appealing.

For all these reasons, Document Understanding (DU) models need to be able to adapt to different domains, not only based on the document image but also on the internal relationships between objects in different document domains. This adaptation should be present during training, but also during inference, when models may be given documents from scarce domains that have never or barely been seen during training in zero- and few-shot settings.

In recent years, Vision-Language Models (VLMs) have shown remarkable performance in many very challenging image understanding tasks \cite{romero25food}, including DocVQA~\cite{raja2023icdar}, where they are becoming the standard. One of their main strengths is their ability to align visual and textual representations to produce better and more reliable answers. However, VLMs tend to rely too much on visual modality, which can lead to errors, especially when dealing with complex layouts such as slides or infographics, where VLMs cannot fully understand the visual relations between objects~\cite{chen2025towards}.

Despite the advances made in the field of DocVQA, real-world deployment requires more robustness across different document domains, which current VLMs still lack. However, there are few comprehensive studies on how pretrained VLMs handle different VQA domains and, more precisely, how they handle different document domains in DocVQA. Furthermore, finetuning VLMs in domain-specific DocVQA datasets can affect performance on other datasets, so it is relevant to study the difference in VLMs in cross-domain evaluations, both in a zero-shot setting where finetuned VLMs do not see any target domain sample during training, and in a few-shot setting, where a small number of samples from the target domain are seen during training to improve generalization \cite{mancera2026inlora}.

To address this gap, we present this study that assesses the performance of VLMs in DocVQA in general document domains. We selected 8 open-source VLMs and three DocVQA datasets that cover domains with several differences in the question types, the image content and granularity, and the layout structure. We selected VLMs with different numbers of parameters to also assess the effect of parameter scale on DocVQA. We evaluated the models in a zero-shot setting in all the datasets. We also finetune each model on every dataset and provide an in-domain and a cross-domain evaluation, highlighting the difference in performance between settings. Finally, we study the effect of few-shot learning in different domains and how the performance of the models changes when dealing with a small number of training samples from the target domain.

The purpose of this work is to provide researchers and practitioners with a comparative study of small-scale VLMs on different DocVQA tasks, highlighting their performance in different settings and testing their adaptability using usual domain adaptation \cite{mancera2026auditing} and knowledge transfer techniques. This study serves as an initial comprehensive evaluation framework for new research and methodologies on automatic domain adaptation and transferability.

The main contributions of this work are:

\begin{itemize}
    \item We provide a comprehensive evaluation of the capabilities of 8 pretrained open-source VLMs on DocVQA in 3 different document domains, covering industrial documents of different types, infographics, and presentation slides. We study which of the models performs better in every domain and how the number of parameters affects the performance.
    \item We separately finetune each model on all the selected domains. We provide a comprehensive in-domain evaluation of each finetuned model, as well as a cross-domain evaluation of all the models in all domains. We show the difference in performance in all settings compared with the zero-shot evaluation of the models. 
    \item We provide an evaluation of the effect of few-shot learning on finetuned VLMs in a cross-domain setting. We study how the number of cross-domain samples affects the VLM performance and the difference with a fully supervised finetuning on the target domain.
\end{itemize}

The rest of the work is structured as follows: Section \ref{sec::related works} provides a study of previous works on VLMs for DU, DocVQA, and domain adaptation techniques in DU. In Section \ref{sec::experimental setup} we describe the models and datasets used for our experiments, and in Section~\ref{sec::experiments and results} we report the results of the zero-shot, supervised finetuning, and few-shot experiments and discuss them. Finally, the conclusions and future research directions emerging from this work are drawn in Section~\ref{sec::conclusions and future work}.

\section{Related Works}
\label{sec::related works}

\subsection{Vision Language Models in Document Understanding}

Document understanding (DU) deals with the interpretation and comprehension of digital or scanned documents, including forms, tables, academic research papers, or slides. There are two broad categories of DU techniques. The first category uses Optical Character Recognition (OCR) annotations provided by external tools~\cite{hong2022bros,xu2021layoutlmv2}. The second category does not use external OCR tools~\cite{kim2022ocr}. 

In DU tasks, VLMs have shown remarkable performance using their own visual encoders enhanced with external OCR tools. LayoutLLM~\cite{luo2024layoutllm} leverages layout instruction tuning in multiple levels of granularity during pretraining and supervised finetuning stages using pretrained weights from LayoutLMv3~\cite{huang2022layoutlmv3}. Other works combine layout information with a contrastive learning environment to enhance visual understanding of VLMs~\cite{li2024enhancing}. Although most of the works leverage the layout information for DU with VLMs, some works have also used VLMs to extract key information from historical documents without OCR tools~\cite{vafaie2025end}.

\subsection{Document Visual Question Answering}

With the development of more sophisticated techniques, VQA has expanded to different and more complex domains such as general document images. DocVQA has received great attention from researchers and practitioners since its establishment. One of the main reasons is that the datasets for DocVQA are quite large and cover a wide range of document domains and question types~\cite{mathew2021docvqa,tanaka2023slidevqa}. Another reason is a very active DocVQA research community, which has led to the organization of very challenging competitions such as the ICDAR 2021 competition in DocVQA~\cite{tito2021icdar} or the ICDAR 2023 competition in VQA in business documents~\cite{raja2023icdar}. In this environment, several techniques have shown impressive performance in DocVQA tasks. For example, ORCA~\cite{lassoued2026orca} uses a specialized multi-agent framework in several stages to exploit the full capabilities of VLMs in DocVQA in a collaborative fashion. Other works perform a contrastive and generative pretraining to align visual encoders and LLMs in general NLP tasks, which have concretely shown impressive performance on DocVQA tasks~\cite{chen2024internvl}. 

\subsection{Domain Adaptation Techniques for Document Understanding}

In real-world applications, it is necessary for deployable models to deal with samples from domains that the model has never seen during training \cite{mancera2025dai}. Domain adaptation ideally reduces this gap by increasing the adaptability of models to new samples while maintaining the same performance on previous samples. In DU tasks, several domain adaptation techniques have been proposed. One of the most prominent approaches is few-shot learning, especially in low-resource domains, where models are required to learn informative representations on target data with few samples for training. For example, FS-DAG~\cite{agarwal2025fs} combines different few-shot learning strategies, positional embeddings, and Graph Neural Networks \cite{lopez2025benchmarking} for Key Information Extraction. Another well-established domain adaptation technique is the generation of synthetic data \cite{bioengineering13050511}, where, given a small source dataset, new data for the same task are generated based on the distribution of the original dataset. In DU, synthetic data have been used for a wide range of sub-tasks such as Financial QA~\cite{harsha2025synthetic}, generalistic QA~\cite{khan2023q} or Table QA~\cite{jiang2022omnitab}.

\section{Experimental Setup}
\label{sec::experimental setup}


\subsection{VLM Selection}

We selected 8 different VLMs from 3 families of models for the experiments in this study: Qwen3.5, Qwen3-VL, and Gemma 3. 

Alibaba's Qwen family~\footnote{\url{https://huggingface.co/qwen}} is an open-weight family of Large Multimodal Language Models (LMLMs) that are showing excellent results on different DU tasks, like ICDAR's 2021 and 2023 competitions on DocVQA~\cite{tito2021icdar,raja2023icdar}. 

The Qwen3-VL sub-family~\cite{bai2025qwen3} is a collection of VLMs designed to handle dynamic-resolution visual input \cite{alcala25attzoom} and process combinations of text and images within a single prompt. The main strengths of the Qwen3-VL family is its scalable performance, showing robust performance in VQA tasks ranging from its tiny 2B version up to its bigger versions, and its long-context multimodality. Due to hardware constraints, especially when finetuning the models, we selected the 2B, 4B, and 8B versions of the Qwen3-VL family for our experiments.

After developing the Qwen3-VL family, Alibaba released a better family of omnimodal models and VLMs, the Qwen3.5 family~\cite{team2026qwen3}. Although originally designed for agentic applications, they also show impressive performance, as usual, for various tasks. One of their main strengths is their efficiency and speed, as they use a Mixture of Experts (MoE) \cite{2018_INFFUS_MCSreview1_Fierrez} and a highly efficient pipeline that cuts the memory overhead without degrading the models performance. Another major advantage of this family is that it is heavily biased towards analytical and structural tasks due to its agentic focus. Although this feature makes them worse in conversational tasks, it makes them especially strong in tasks like VQA, where the models are required to just output the answer of the given question. In this case, to show the different performance between large scale and smaller models, we selected the 0.8B, 2B, 4B and 9B versions of this family for our experiments.

Google DeepMind's Gemma 3~\footnote{\url{https://huggingface.co/collections/google/gemma-3-release}} is a family of lightweight open multimodal models specially designed for visual and textual understanding and optimized for developer accesibility. Their main strength is that they are specially receptive to parameter-efficient finetuning, which makes them excellent at adapting to new tasks never seen during their pretraining, as is the case of DocVQA. However, they tend to hallucinate when adapting to new tasks when the knowledge is not effectively embedded during finetuning. Due to hardware limitations, we only selected the Gemma 3 4B version of this family for our experiments.

All models were implemented and finetuned using the Unsloth library~\footnote{\url{https://github.com/unslothai/unsloth}}, which is a library for efficient inference and finetuning of LLMs and VLMs.

\subsection{Datasets}

To reflect the real-world diversity of documents, we selected three datasets covering well-differentiated domains: 1) Single Page DocVQA (SP-DocVQA), which covers industrial documents of various types; 2) InfographicsVQA, made up of infographics; and 3) SlideVQA, a dataset comprising presentation slides.

SP-DocVQA and InfographicsVQA are single-page DocVQA datasets, that is, the input to the models is a unique document image and question. The answer of that question must be present in the document image or be extracted from it. In contrast, SlideVQA is designed as a multi-page DocVQA dataset, that is, the models are given various document images, in this case 20 presentation slides, and a question and models are required to answer the given question by retrieving the necessary images from the images and answering the question based on the context of multiple images. To keep a fair comparison between datasets, we converted SlideVQA dataset into a single page DocVQA dataset by selecting a subset of it, as we will explain in subsequent sections. 

\subsubsection{SP-DocVQA:}

\begin{figure}[t]
    \centering
    \includegraphics[width=\textwidth]{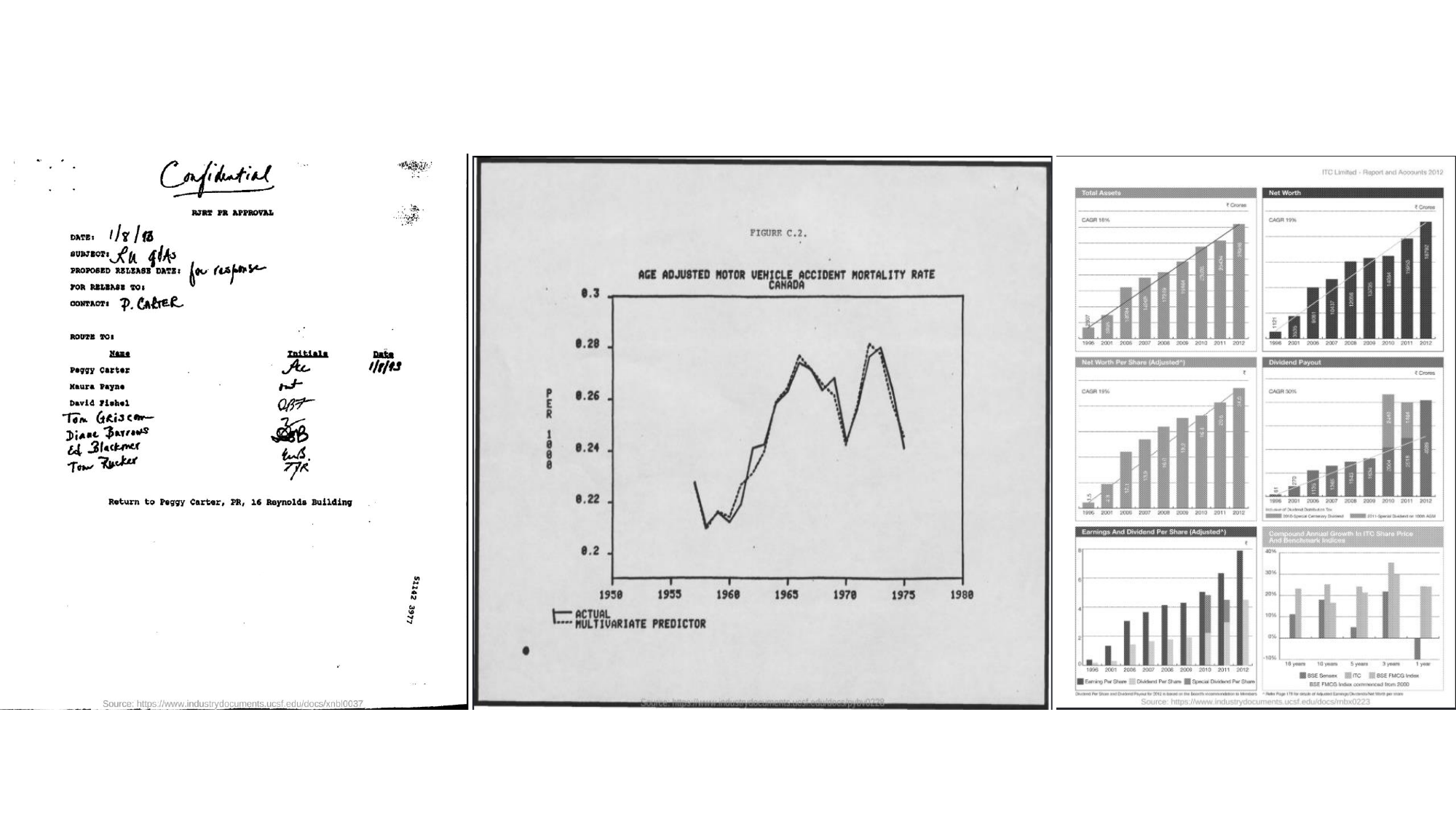}
    \caption{Samples from SP-DocVQA~\cite{mathew2021docvqa}. Different documents present different layouts, but most of them present an stable structure and reading order.}
    \label{fig:combined_sp_docvqa}
\end{figure}

Presented in 2021, the SP-DocVQA dataset~\cite{mathew2021docvqa} is one of the most representative DocVQA datasets in the literature. It comprises 50K Question-Answer pairs over more than 12K document images. The document images were collected from documents in the UCSF Industry Documents Library~\footnote{\url{https://www.industrydocuments.ucsf.edu/}}. The documents in the dataset were selected to cover $5$ industries: Tobacco, Food, Drug, Fossil Fuel, and Chemical. The selected documents range from 1900 to 2018. 

SP-DocVQA covers a wide range of document types and layouts, as can be seen in Fig.~\ref{fig:combined_sp_docvqa}. Most of the documents are letters, forms, and reports, representing approximately 50\% of the documents in the dataset, but there are other types of document, covering an even more general domain. 

\subsubsection{InfographicsVQA:}

\begin{figure}[t]
    \centering
    \includegraphics[width=\textwidth]{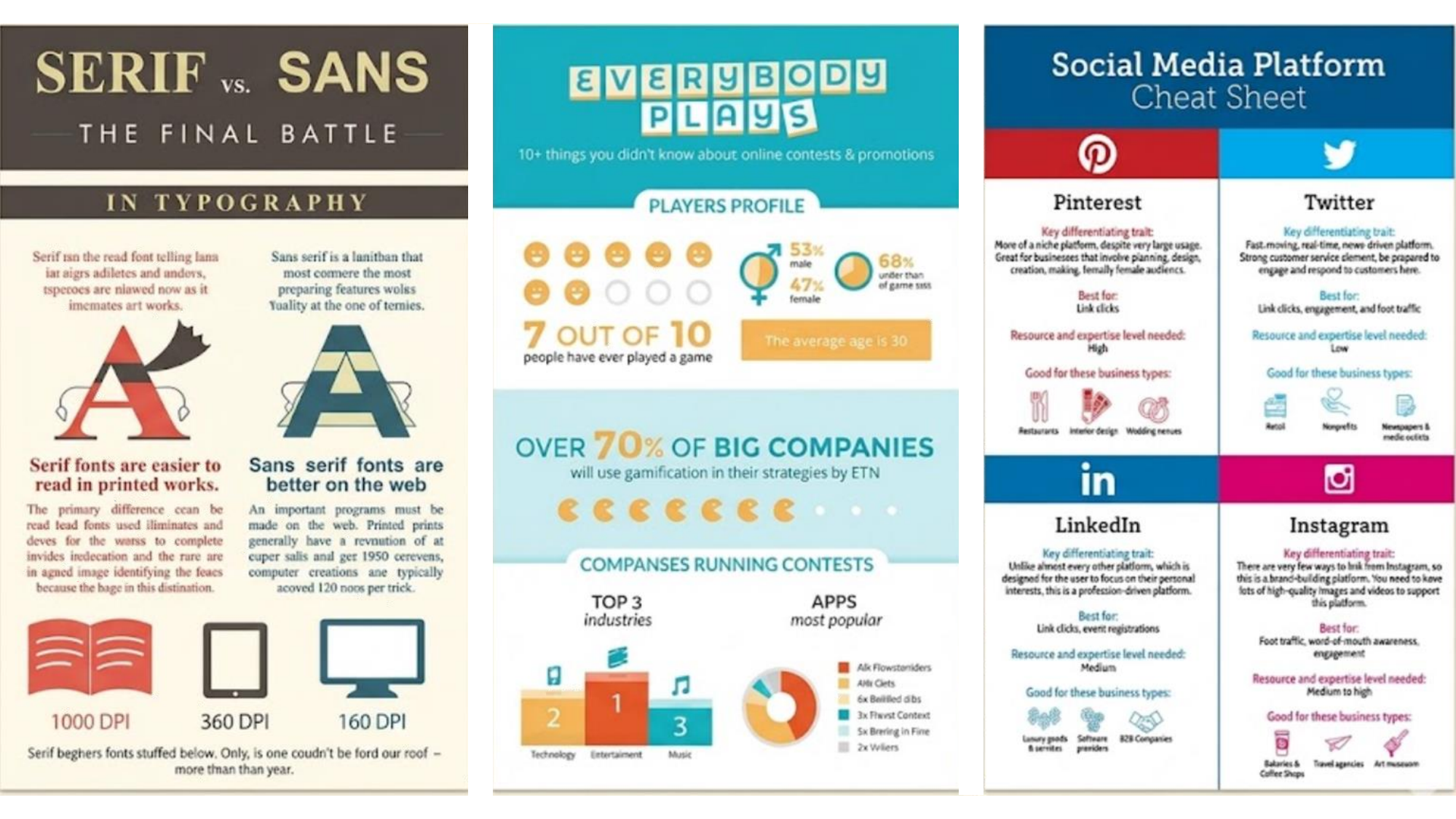}
    \caption{Samples from InfographicsVQA~\cite{mathew2022infographicvqa}. (Some of them were cropped.)}
    \label{fig:combined_infographics}
\end{figure}

InfographicsVQA~\cite{mathew2022infographicvqa} was presented in 2022 to benchmark the progress of VQA research combining vision, language, and document understanding. It comprises approximately 30K question-answer pairs for more than 5K document images. These document images come from infographics collected from the Internet and later curated. The questions in InfographicsVQA are divided into 4 categories: 1) Image-span, where the answer to the question is extracted from a token or a sequence or tokens present in the image; 2) Question-span, where the answer is inside the question itself; 3) Multi-span, where the answer is present in the image in multiple spans; and 4) Non-extractive, where the answers are numerical and not present in the document image itself. 

Contrary to SP-DocVQA, where all questions are extractive, InfographicsVQA presents a more complex task as there are non-extractive questions, which force VQA models to reason over the image and not only extract tokens from it. Another major difference from SP-DocVQA is the kind of layouts present in InfographicsVQA. The infographics show more complex layouts, as shown in Fig.~\ref{fig:combined_infographics}, where the reading order and the relationship between objects are not easily inferred and may be subjective. This adds another layer of complexity to the VQA task in this domain.

\subsubsection{SlideVQA:}

\begin{figure}[t]
    \centering
    \includegraphics[width=\textwidth]{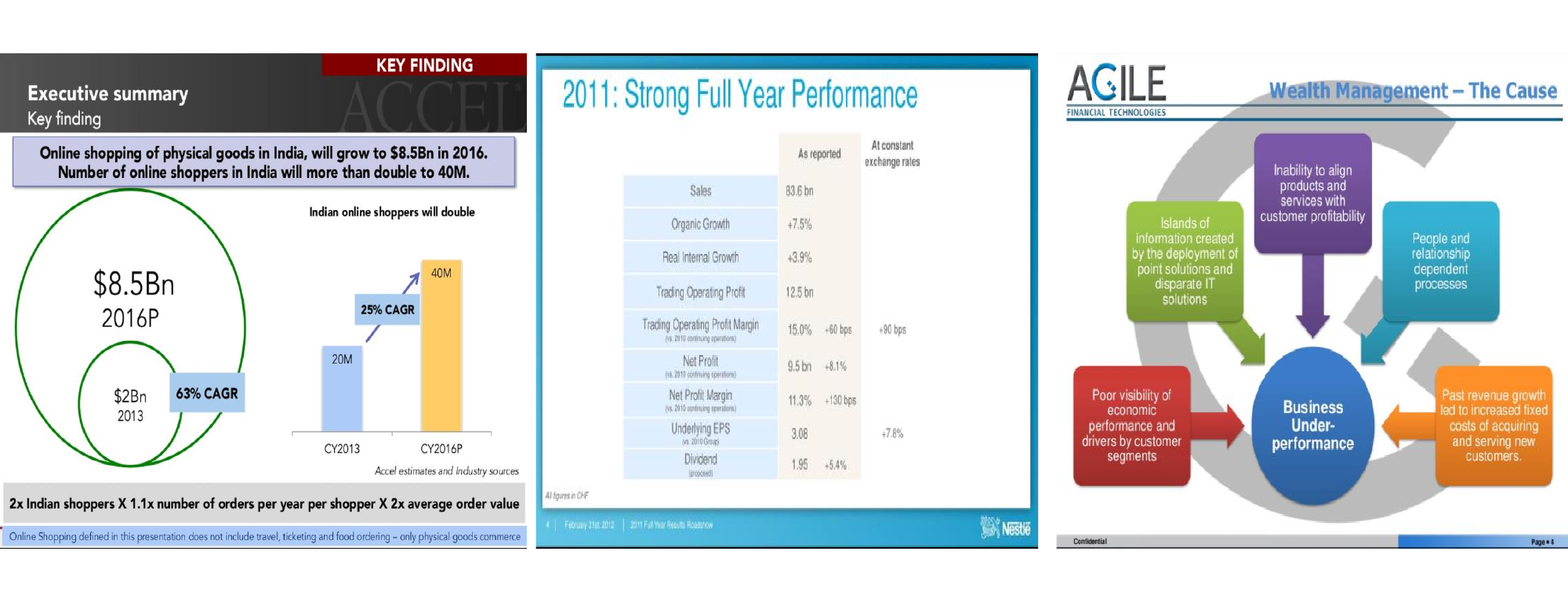}
    \caption{Samples from SlideVQA~\cite{tanaka2023slidevqa}. Different slides present big differences in layout structure and visual artifacts.}
    \label{fig:combined_slidevqa}
\end{figure}

In 2023, SlideVQA~\cite{tanaka2023slidevqa} was presented in the Thirty-Seventh AAAI Conference on Artificial Intelligence (AAAI-23). This dataset comprises more than 14K question-answer pairs from around 2.5K slide decks. The questions in this dataset are categorized into three types: 1) Single- and 2) Multi-span, which are questions whose answers are extracted from a single or various spans respectively of one or more evidence pages; and 3) Non-span questions, which are questions whose answers are composed of numerical values or visual appearances. The questions in SlideVQA are further divided in single-hop or multi-hop, depending on the number of slide pages from where the information need to be extracted to answer the question. To keep a fair comparison with the other selected datasets, which are single-page DocVQA datasets, we filtered out the multi-hop questions in SlideVQA, and kept only the single-hop questions. 

Similarly to InfographicsVQA, the questions in SlideVQA are not only extractive, but also numerical and based on visual cues. This presents a complex task for models, as they not only need to reason over tokens, but also show mathematical reasoning to answer the question correctly. Another important feature of SlideVQA is that the slides do not show a regular or structured layout, as shown in Fig.~\ref{fig:combined_slidevqa}, not even slides from the same presentation. Slides present less text than infographics or forms, so the relationships between objects are mostly based on visually-grounded information, and not so text-dependent. This presents a key factor when working with VLMs on slides, as the visual encoder of the model takes much more importance than the text encoder, which is also one of the reasons why VLMs may fail to adapt to domains where the importance between vision and language is not balanced and is variable between domains.

\section{Experiments and Results}
\label{sec::experiments and results}


\subsection{Zero-Shot Experiments}

\begin{table}[t]
\caption{Mean ANLS in a \textbf{Zero-shot setting} for each model family and selected number of parameters on the evaluation set of each DocVQA dataset.} 
\centering
\setlength{\tabcolsep}{8pt} 
\begin{tabular}{@{}lcccc@{}}
\toprule
\multirow{2}{*}{\textbf{Model}} & \multirow{2}{*}{\textbf{\# Params}} & \multicolumn{3}{c}{\textbf{Test Dataset (ANLS $\uparrow$)}} \\ \cmidrule(l){3-5} 
 & & \textbf{SP-DocVQA} & \textbf{InfographicsVQA} & \textbf{SlideVQA} \\ \midrule
\multirow{3}{*}{Qwen3-VL}            & 2B              & $85,36$          & $44,78$         & $40,33$         \\
& 4B              & $89,87$          & $54,27$         & $46,17$         \\
                 &  8B   & $91,98$          & $57,55$         & $49,63$         \\ \hline
\multirow{4}{*}{Qwen3.5}            & 0.8B              & $71.9$          & $30.63$         & $22.01$         \\
& 2B              & $84,96$          & $41,21$         & $36,05$         \\
& 4B              & $90,86$          & $54,21$         & $46,76$         \\
                 &  9B   & $\mathbf{92,67}$          & $\mathbf{60,06}$         & $\mathbf{51,38}$         \\ \hline
\multirow{1}{*}{Gemma 3}            & 4B              & $73,61$          & $37,96$         & $49,04$         \\ \hline
\end{tabular}
\label{tab::zero shot}
\end{table}

The first experiments we ran consisted of running a zero-shot evaluation of the 8 selected VLMs in the three datasets. Before passing the document images through the model, we padded them to unify the image sizes of all the datasets. The final image resolution of all images was $1449 \times 1449$. We found that padding the images to a unified size yields better performance for all settings than passing the original images of different sizes and resolutions through the models. Apart from the image, we also passed the question at hand with the following prompt: \\
\texttt{"Answer the following question, show only the answer of the question. \\
Question: \\
Answer:"}.

The results of the zero-shot evaluation are reported in Table~\ref{tab::zero shot}. We report the mean Average Normalized Levenshtein Similarity (ANLS) over the evaluation set of each dataset. As expected, the best performing VLMs are the ones with more number of parameters, concretely Qwen3-VL 8B and Qwen 3.5 9B. It is notable that for all three datasets, the VLMs performance increases with the number of parameters, which reflects the complexity of the DocVQA task, as smaller models lack the necessary depth of understanding to fully exploit the necessary visual and text artifacts to answer the questions correctly, at least in a zero-shot setting. It is also remarkable the great difference in ANLS performance of each VLM between datasets. In SP-DocVQA, all VLM show impressive performance in a zero-shot environment that surpasses in some cases the $90\%$ of ANLS, while in InfographicsVQA and SlideVQA the best performing VLM barely surpasses the $60\%$ and $50\%$, respectively. We believe that this difference is due to the visual and layout complexity of InfographicsVQA and SlideVQA, which pretrained VLMs without any finetuning are not able to fully comprehend, while the structured layout of SP-DocVQA lets VLMs to fully understand the documents, and leverage the internal previous knowledge from pretraining to understand and correctly answer the questions.

\subsection{Supervised Finetuning}

Previous works in VQA have addressed the possibility that VLMs cannot fully exploit their internal knowledge when answering questions~\cite{khan2023q}. To test that hypothesis in DocVQA, we decided to perform a fully supervised finetuning of each model on every dataset. This would allow us to determine if the bottlenecks when dealing with document images are the questions or the visual complexity.

For this experiment, all models were finetuned using the Unsloth library for $3$ epochs using LoRA~\cite{hu2022lora,mancera2026auditing,mancera2026inlora} with rank $32$, LoRA alpha $4$ and LoRA dropout of $0.2$. For each finetuning, we report the results of inter- and intra-dataset evaluations. 

\subsubsection{Finetuning on SP-DocVQA.}
\begin{table}[t]
\centering
\caption{Mean ANLS results for VLMs finetuned on \textbf{SP-DocVQA}. We report the difference in ANLS with respect to the Zero-shot setting (Table~\ref{tab::zero shot}).\\}
\label{tab:finetune_spdocvqa}
\setlength{\tabcolsep}{8pt}
\begin{tabular}{@{}lcccc@{}}
\toprule
\multirow{2}{*}{\textbf{Model}} & \multirow{2}{*}{\textbf{\# Params}} & \multicolumn{3}{c}{\textbf{Test Dataset (ANLS $\uparrow$)}} \\ \cmidrule(l){3-5} 
 & & \textbf{SP-DocVQA} & \textbf{InfographicsVQA} & \textbf{SlideVQA} \\ \midrule
\multirow{3}{*}{Qwen3-VL} & 2B & $91.63 \;(+\, 6.27)$ & $57.38 \;(+\, 12.60)$ & $44.70 \;(+\, 4.37)$ \\
 & 4B & $94.09 \;(+\, 4.22)$ & $66.72 \; (+ \, 12.45)$ & $50.13 \;(+\, 3.96)$ \\
 & 8B & $95.00 \; (+ \, 3.02)$  & $71.15 \; (+ \, 13.60)$ & $51.69 \;(+\, 2.06)$ \\ \midrule
\multirow{4}{*}{Qwen3.5} & 0.8B & $89.35 \; (+\, 17.45)$ & $48.00\; (+\, 17.37)$ & $35.54 \; (+ \, 13.53)$ \\
 & 2B & $93.42 \; (+\, 8.46)$ & $61.38\; (+\, 20.17)$ & $46.46 \; (+ \, 10.41)$ \\
 & 4B & $95.18 \; (+ \, 4.32)$ & $68.89 \; (+ \, 14.68)$ & $50.89 \; (+ \, 4.13)$ \\
 & 9B & $96.55 \; (+ \, 3.88)$ & $74.26 \; (+ \, 14.20)$ & $54.51 \; (+ \, 3.13)$ \\ \midrule
Gemma 3 & 4B & $81.11 \; (+ \, 7.5)$ & $35.72 \; (\red{\mathbf{- \, 2.24}})$ & $50.14 \, (+ \, 1.10)$ \\ \bottomrule
\end{tabular}%
\end{table}
In Table~\ref{tab:finetune_spdocvqa} we observe that almost all models outperform the zero-shot setting. As expected, the models that increase more with respect to the zero-shot setting are the smaller ones, concretely, the Qwen 3.5 0.8B and 2B versions. This may confirm our statement that smaller pretrained models lack enough depth to fully understand the DocVQA task in zero-shot evaluations and need some kind of finetuning or context learning to exploit its full potential. The only finetuned model that did not outperform the zero-shot evaluation was Gemma 3 4B, which gave a mean ANLS decrease of $2.24$ in InfographicsVQA. This model is also the one with the worst performance in all the datasets over models with more than 4 billion parameters.

After finetuning SP-DocVQA, we observe that while the largest increase in performance occurs with Qwen 3.5 0.8B in the SP-DocVQA evaluation set, with an increase of 17.45\% in ANLS, there is not much difference between the evaluation on SP-DocVQA and SlideVQA, with an increase in ANLS of approximately the same percentages. This is not the case on InfographicsVQA, where the finetuning on SP-DocVQA gave the biggest increase in ANLS with respect to the zero-shot setting among the three datasets.   

\subsubsection{Finetuning on InfographicsVQA.}
\begin{table}[p]
\centering
\caption{Mean ANLS results for VLMs finetuned on \textbf{InfographicsVQA}. We report the difference in ANLS with respect to the Zero-shot setting (Table~\ref{tab::zero shot}).\\}
\label{tab:finetune_infographicsvqa}
\setlength{\tabcolsep}{8pt}
\begin{tabular}{@{}lcccc@{}}
\toprule
\multirow{2}{*}{\textbf{Model}} & \multirow{2}{*}{\textbf{\# Params}} & \multicolumn{3}{c}{\textbf{Test Dataset (ANLS $\uparrow$)}} \\ \cmidrule(l){3-5} 
 & & \textbf{SP-DocVQA} & \textbf{InfographicsVQA} & \textbf{SlideVQA} \\ \midrule
\multirow{3}{*}{Qwen3-VL} & 2B & $89.49 \; (+ \, 4.13)$ & $60.37 \; (+ \, 15.59)$ & $45.74 \; (+ \, 5.41)$ \\
 & 4B & $93.26 \; (+ \, 3.39)$ & $69.83 \; (+ \, 15.56)$ & $50.60 \; (+ 4.43)$ \\
 & 8B & $93.80 \; (+ \, 1.82)$ & $73.23 \; (+ \, 15.68)$ & $52.96 \; (+ \, 3.33)$ \\ \midrule
\multirow{4}{*}{Qwen3.5} & 0.8B & $85.76 \; (+ \, 13.86)$ & $53.95 \; (+ \, 23.32)$ & $38.32 \; (+ \, 16.31)$ \\
 & 2B & $92.08 \; (+ \, 7.12)$ & $66.24 \; (+ \, 25.03)$ & $48.50 \; (+ \, 12.45)$ \\
 & 4B & $94.65 \; (+ \, 3.79)$ & $74.00 \; (+ \, 19.79)$ & $52.00 \; (+ \, 5.24)$ \\
 & 9B & $95.36 \; (+ \, 2.69)$ & $78.68 \; (+ \, 18.62)$ & $55.21 \; (+ \, 3.83)$ \\ \midrule
Gemma 3 & 4B & $75.02 \; (+ \, 1.41)$ & $43.68 \; (+ \, 5.72)$ & $55.75 \; (+ \, 6.71)$ \\ \bottomrule
\end{tabular}%
\end{table}

\begin{table}[p]
\centering
\caption{Mean ANLS results for VLMs finetuned on \textbf{SlideVQA}. We report the difference in ANLS with respect to the Zero-shot setting (Table~\ref{tab::zero shot}).\\}
\label{tab:finetune_slidevqa}
\setlength{\tabcolsep}{8pt}
\begin{tabular}{@{}lcccc@{}}
\toprule
\multirow{2}{*}{\textbf{Model}} & \multirow{2}{*}{\textbf{\# Params}} & \multicolumn{3}{c}{\textbf{Test Dataset (ANLS $\uparrow$)}} \\ \cmidrule(l){3-5} 
 & & \textbf{SP-DocVQA} & \textbf{InfographicsVQA} & \textbf{SlideVQA} \\ \midrule
\multirow{3}{*}{Qwen3-VL} & 2B & $89.54 \; (+ \, 4.18)$ & $55.78 \; (+ \, 11.00)$ & $48.47 \; (+ \, 8.14)$ \\
 & 4B & $92.38 \; (+ \, 2.51)$ & $64.75 \; (+ \, 10.48)$ & $50.92 \; (+ \, 4.75)$ \\
 & 8B & $92.89 \; (+ \, 0.91)$ & $70.85 \; (+ 13.30)$ & $57.90 \; (+ \, 8.27)$ \\ \midrule
\multirow{4}{*}{Qwen3.5} & 0.8B & $85.59 \; (+ \, 13.69)$ & $46.05 \; (+ \, 15.42)$ & $40.05 \; (+ \, 18.04)$ \\
 & 2B & $91.47 \; (+ \, 6.51)$ & $60.62 \; (+ \, 19.41)$ & $50.77 \; (+ \, 14.72)$ \\
 & 4B & $93.46 \; (+ \, 2.60)$ & $68.49 \; (+ \, 14.28)$ & $53.97 \; (+ \, 7.21)$ \\
 & 9B & $93.17 \; (+ \, 0.50)$ & $73.51 \; (+ \, 13.45)$ & $59.65 \; (+ \, 8.27)$ \\ \midrule
Gemma 3 & 4B & $73.31 \; (\red{\mathbf{-\, 0.30}})$ & $36.43 \; (\red{\mathbf{- \, 1.53}})$ & $58.99 \; (+ \, 9.95)$ \\ \bottomrule
\end{tabular}%
\end{table}

In Table~\ref{tab:finetune_infographicsvqa} we observe a similar performance in both the inter- and intra-data evaluations after finetuning in InfographicsVQA and after finetuning in SP-DocVQA. As expected, the inter-dataset evaluation gives better results than the inter-dataset evaluation with SP-DocVQA. In particular, the biggest differences of these results with respect to the results in Table~\ref{tab:finetune_spdocvqa} are the Gemma 3 results in InfographicsVQA and SlideVQA, with an increase of $5.72 \%$ and $6.71 \%$ respectively with respect to the zero-shot evaluation, in contrast to the results after finetuning in SP-DocVQA, where there was a decrease in ANLS when evaluating in InfographicsVQA. Another remarkable difference is that all models perform better in SlideVQA after being finetuned in InfographicsVQA than after being finetuned in SP-DocVQA, which may indicate the similarity of both datasets in their layouts and visual complexity. These results show that VLMs need to address different layout structures more efficiently to extract key visual information from different document types.

\subsubsection{Finetuning on SlideVQA.}

Finally, in Table~\ref{tab:finetune_slidevqa} we observe similar results between and within the data set to those reported earlier. However, after finetuning on SlideVQA we noticed that the increase in ANLS on inter-dataset evaluations is smaller than when finetuning on SP-DocVQA and InfographicsVQA. This difference is more notable with larger models such as Qwen 3.5 9B and Qwen 3 VL 8B, where the difference in ANLS in SP-DocVQA with respect to zero-shot evaluation is less than $1 \%$. It is also notable that Gemma 3 4B is the only model that performs worse on SP-DocVQA after being finetuned for the DocVQA task. Concretely, after finetuning Gemma 3 4B in SlideVQA it gives a decrease in the ANLS between data sets of $0.30 \%$ in SP-DocVQA.

\subsection{Few-Shot Results}


\begin{figure}[p]
    \centering
    \begin{subfigure}{\textwidth}
        \centering
        \includegraphics[width=\linewidth]{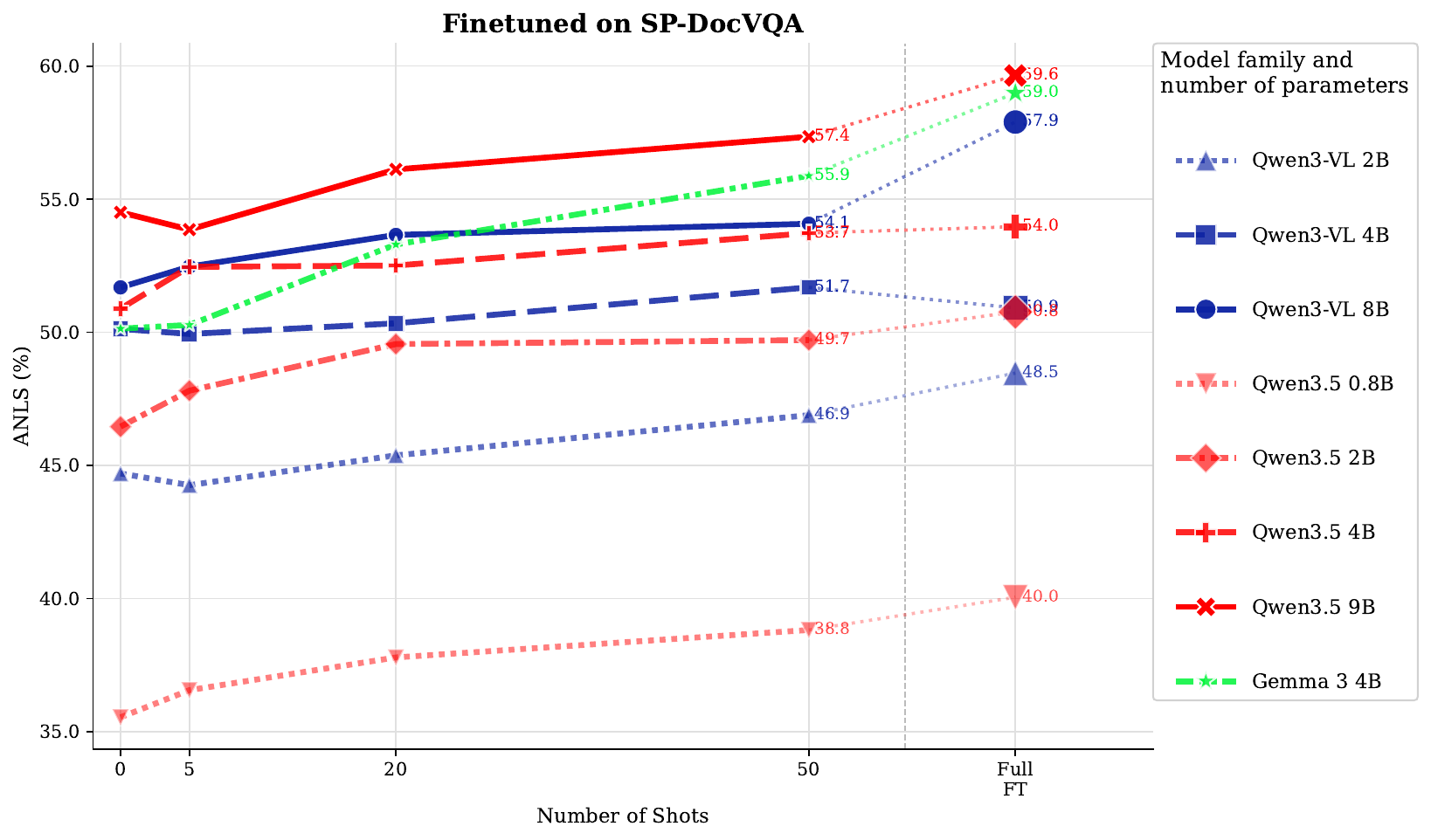}
        \caption{}
        \label{fig:fewshot_spdocvqa}
    \end{subfigure}
    
    \vspace{5mm} 
    
    \begin{subfigure}{\textwidth}
        \centering
        \includegraphics[width=\linewidth]{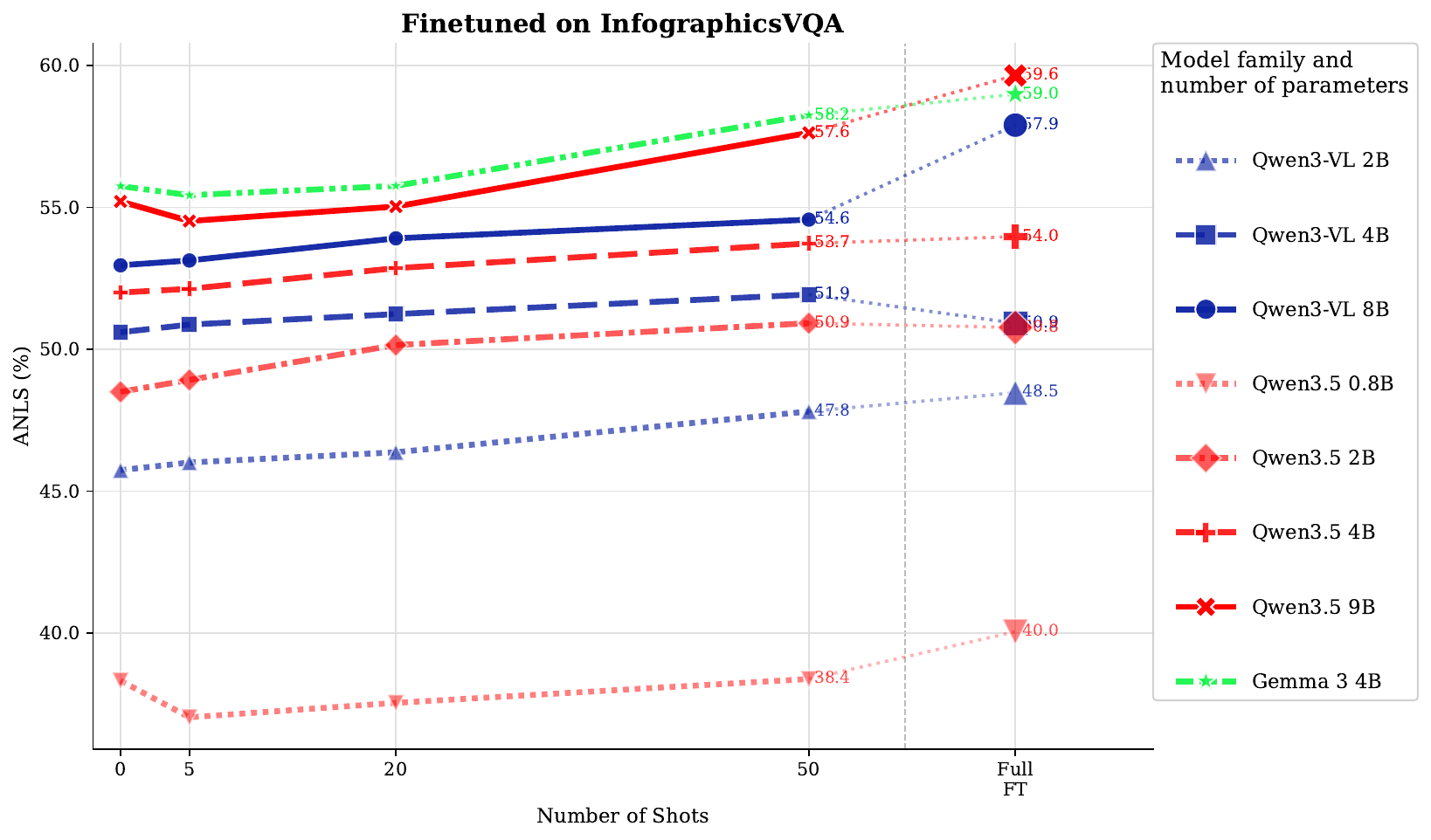}
        \caption{}
        \label{fig:fewshot_infovqa}
    \end{subfigure}
    
    \caption{Mean ANLS on SlideVQA of finetuned models on SP-DocVQA (a) and InfographicsVQA (b) for different number of shots. Zero-shot results correspond to the SP-DocVQA and InfographicsVQA finetuning cross-dataset evaluation on SlideVQA. Full finetuning (Full FT) results correspond to the intra-dataset results after finetuning on SlideVQA.}
    \label{fig:fewshot_results}
\end{figure}

In order to provide a more complete study on the effect of finetuning VLMs on different document domains, we performed a final round of few-shot experiments. For this round of experiments, we selected SlideVQA as the target dataset, as it is the dataset where most models perform worse even after specific finetuning on SlideVQA, and SP-DocVQA and InfographicsVQA as the source datasets. 

We selected the finetuned models on SP-DocVQA and InfographicsVQA and finetuned them for 3 more epochs using a small number of training samples. The number of SlideVQA samples on which the model was trained was $5, 20$ and $50$ samples. Fig.~\ref{fig:fewshot_results} reports a graphic showing the evolution of the results of each VLM with increasing shots. We compare the performance of few-shots with the zero-shot evaluation of each finetuned model in SP-DocVQA based on the results reported in Table~\ref{tab:finetune_spdocvqa} and  of each finetuned model in InfographicsVQA based on the results reported in Table~\ref{tab:finetune_infographicsvqa}. We also compare the few-shot results with the ``best'' possible results, which are expected to be the results after a fully supervised finetuning on SlideVQA.

We observe that most models perform better when more training samples are added, with some exceptions such as Qwen3.5 9B, which performed worse with $5$ samples than with $0$ samples. As expected, all models perform better with $50$ training samples than with zero samples, which are the zero-shot results. However, we observe that, in general, the difference between full finetuning and finetuning using only $50$ samples is not that much, and some VLMs even perform better when given only $50$ training samples than with the entire training set. 

These results may lead to the conclusion that pretrained and finetuned VLMs are already good at answering questions based on visual inputs, which can be inferred from the results discussed in previous sections. However, there is a huge bottleneck for VLMs, as their performance is greatly constrained by an inefficient extraction of visual information and, more concretely, a suboptimal understanding of complex layouts like the ones present on InfographicsVQA and SlideVQA.

\section{Discussion and Future Work}
\label{sec::conclusions and future work}

In this work, we presented a comprehensive empirical study evaluating the capabilities, robustness and ability to adapt to different domains of $8$ Vision-Language Models on distinct Document Visual Question Answering domains. We selected datasets $3$ from DocVQA consisting of industrial documents of a wide range of document types, infographics, and presentation slides. We studied how the VLM family, the number of parameters and the cross-domain training influence the DocVQA task. Our experiments cover zero-shot evaluations, fully supervised finetuning with inter- and intra-dataset evaluations, and few-shot finetuning on a target domain of already finetuned models on different domains. All of these experiments provide a comprehensive understanding of the strengths and weaknesses of VLMs in different document domains for DocVQA. This may serve for future work on domain adaptation techniques using VLMs in DocVQA.

Our experiments reveal several insights regarding layout and visual complexity when working with VLMs. In zero-shot settings, pretrained models show strong performance in highly structured and easy-to-read layouts like the ones on SP-DocVQA, where an ANLS of more than $90\%$ is achieved. However, this performance dropped sharply when confronted with complex layouts and visually dependent structures of infographics and presentation slides, which may reveal that the usual text-vision alignment of VLMs is not enough to fully exploit the models pretrained knowledge without a correct structural or domain-dependent adaptation. Furthermore, while parameter scaling consistently proved to be an important factor in performance, smaller models exhibit the highest relative gains during fully supervised finetuning. This shows that supervised optimization allows smaller architectures to demonstrate their full capabilities. We also observed that similarities in layout complexity plays an important role in generalization, as models trained on visually complex infographics transferred their knowledge to presentation slides more effectively than those trained on structured layouts present in industrial documents.

Few-shot experiments on presentation slides provide an encouraging outlook for low-resource domains in real-world scenarios. Incorporating as few as $50$ target-domain samples into previously finetuned models on large-scale datasets led to performance improvements, with some models outperforming their fully supervised counterparts. This behavior demonstrates that the underlying bottleneck during cross-domain knowledge transfer is not a lack of domain knowledge, but rather a structural domain misalignment.

Several research directions emerge from this study. Future works should explore architectural modifications that explicitly inject domain information and modules to provide VLMs with easier-to-exploit representations of visually complex layouts to fully exploit their capacities. Additionally, mitigating negative cross-domain transfer and preventing catastrophic forgetting remain an open challenge, suggesting that parameter-efficient knowledge distillation techniques may be the way to go for researchers. Finally, this study provides a study on single page DocVQA, however, multi-page DocVQA remains a complex challenge, especially in multi-domain scenarios or open-domain settings, and a similar study on multi-page DocVQA would be of great use for the community. 

In our future work, we will also study the risks in LLMs and VLMs both in the user interaction with them~\cite{daza2026ares} and intrinsically based on their nature and learning elements \cite{mancera2026inlora,dealcala2026my}). Analyzing biases \cite{2023_ECAIw_LFIT-XAI_Tello,pena2025addressing,serna27unravel} and synthetic manipulations in documents \cite{pavel25iccv,MUNOZHARO2026103969} while maintaining privacy \cite{2017_Access_HEmultiDTW_Marta,mancera2025pba} are also key topics in our DU research agenda. Finally, we will apply these DU technologies to specific domains such as crime investigation \cite{miguel26crime}, public and legal affairs \cite{PENA-Layout,lopez2025benchmarking}, e-learning~\cite{irigoyen26k12}, and e-health~\cite{ROMEROTAPIADOR2026111676,bioengineering13050511}.

\begin{credits}
\subsubsection{\ackname}
Supported by M2RAI (PID2024-160053OB-I00 MICIU/FEDER), Cátedra ENIA UAM-VERIDAS en IA Responsable (NextGenerationEU PRTR TSI-100927-2023-2), and Research Agreement DGGC/UAM/FUAM for Biometrics and Applied AI. Work conducted within the ELLIS Unit Madrid. Lopez-Duran is supported by a FPI Fellowship (FPI-UAM-2025). Robledo-Moreno is supported by a FPI Fellowship (FPI-UAM-2025). DeAlcala is supported by a FPU Fellowship (FPU21/05785). 
\subsubsection{\discintname} The authors declare no conflict of interests.
\end{credits}

%
%
%
\bibliographystyle{splncs04}
\bibliography{main}
\end{document}